\pdfoutput=1

\documentclass[11pt]{article}

\usepackage[]{latex/acl}

\usepackage{times}
\usepackage{latexsym}

\usepackage[T1]{fontenc}

\usepackage[utf8]{inputenc}

\usepackage{microtype}

\usepackage{inconsolata}

\usepackage{graphicx}
\usepackage{subcaption}
\usepackage[table]{colortbl}
\usepackage{url}
\newcommand*\circled[1]{\raisebox{.5pt}{\textcircled{\raisebox{-1pt} {#1}}}}

\title{\textsc{Attendre}: Wait To Attend By Retrieval With Evicted Queries in Memory-Based Transformers for Long Context Processing}

\author{Zi Yang \\
  Google Research \\
  \texttt{ziy@google.com} \\\And
  Nan Hua \\
  Google Research \\
  \texttt{nhua@google.com} \\}

\begin{document}
\maketitle
\begin{abstract}
As LLMs have become capable of processing more complex types of inputs, researchers have recently studied how to efficiently and affordably process possibly arbitrarily long sequences. One effective approach is to use a FIFO memory to store keys and values of an attention sublayer from past chunks to allow subsequent queries to attend. However, this approach requires a large memory and/or takes into the consideration the specific LM architecture. Moreover, due to the causal nature between the key-values in prior context and the queries at present, this approach cannot be extended to bidirectional attention such as in an encoder-decoder or PrefixLM decoder-only architecture. In this paper, we propose to use eviction policies, such as LRA and LFA, to reduce the memory size and adapt to various architectures, and we also propose the \textsc{Attendre} layer, a wait-to-attend mechanism by retrieving the key-value memory (K/V memory) with evicted queries in the query memory (Q memory). As a first step, we evaluate this method in the context length extension setup using the TriviaQA reading comprehension task, and show the effectiveness of the approach.
\end{abstract}

\section{Introduction}
\label{sec:introduction}

Transformer-based LLMs have become capable of processing more complex types of inputs, including structured documents and multi-modal contents, which calls for efficient and affordable approaches to process possibly arbitrarily long input sequences. Due to the quadratic computational complexity of the dot-product attention in the original Transformer architecture \cite{NIPS2017_3f5ee243}, various \emph{sparsity} and \emph{compression} methods \cite{beltagy2020longformer, zaheer2020big, guo-etal-2022-longt5, phang2022investigating, xiong-etal-2022-simple, ding2023longnet} have been proposed, which reduce the complexity to subquadratic or linear, while still require to read the entire input sequence all at once and truncate tokens beyond a certain context length limit.

To process even longer and possibly arbitrarily long sequences, one eventually has to split the input sequence into chunks and process each chunk at a time. Researcher have proposed approaches that use either a recurrent \emph{state} \cite{gu2021efficiently, ma2022mega, gu2023mamba, lutati-etal-2023-focus, bulatov2023scaling} or a continuous \emph{memory} \cite{dai-etal-2019-transformer, wu2021memorizing, xiao2023efficient} to persist context between the chunks\footnote{The boundary between a recurrent state and a continuous memory may not always be clear. We see a recurrent state undergoes a complete transformation and update at every step, whereas a continuous memory may choose to update partial contents without affecting others.}. Memorizing Transformer \cite{wu2021memorizing} introduces a memory structure to store keys and values of an attention sublayer from past chunks to allow queries in subsequent chunks to attend. In practice, this method often requires a large memory to reach the performance of the original Transformer architecture processing the entire sequence. StreamingLLM \cite{xiao2023efficient} found that a small memory partition that persists only the initial positions can work as well as a full Transformer model. However, this approach is designed specifically for RoPE \cite{su2023roformer} and its variants, as prior works \cite{kazemnejad2023impact, han2023lm} have suggested that the RoPEs of the initial positions contain important absolute position information, and it does not aim to combat the general lost-in-the-middle issue \cite{liu2023lost, peysakhovich2023attention}. Moreover, these approaches usually read the chunks in the temporal order, implying that the queries can only attend the key-values in a prior or present context, not ``future'' context, which makes them impossible to apply to a model pretrained using only or partially bidirectional attention, e.g. the encoder layers in an encoder-decoder or prefix-LM decoder-only architecture.

We aim to tackle these problems and find a memory design that (1) requires a minimum capacity to reach the same level of performance of a FIFO and can be easily adapted to various architectures without reconfiguration, and (2) relaxes the unidirectional restriction of a memory-based method and supports bidirectional attention. We first propose to leverage \emph{cache eviction policies} to memory at insertion time. Inspired by the widely adopted eviction policies LRU (least recently used) and LFU (least frequently used), we introduce a family of LRA (least recently attended) and LFA (least frequently attended) policies, which takes the advantage of the computed attention score and uses it as a proxy for importance, to decide which key-value positions to keep in the memory and which to evict for computation or simply discarding. In our experiment, we find that a memory of size 128 that uses our proposed policy can perform on par with the baseline method that has a size of 2,048. Next, we propose the \textsc{Attendre} layer, a wait-to-\textsc{attend} mechanism by \textsc{r}etrieving the key-value memory (K/V memory) with \textsc{e}victed queries in the query memory (Q memory), where the K/Vs being attended may come from chunks many steps behind the evicted query, i.e. ``future'' chunks from the query perspective. We use the TriviaQA reading comprehension task to evaluate the propose methods, and we find that the methods can help the performance reach the level of the original model processing the entire long sequence.

\section{Related Work}

Long context modeling (or long range modeling, long sequence processing) has become a very broad research topic. In this section, we focus on memory-based Transformer models that support arbitrarily long inputs, in particular, what to memorize and how to update. Other sparsity and compression methods can be orthogonal to and thus combine with ours to achieve better efficiency. Interested readers should refer to other survey papers, e.g. \citet{dong2023survey} and \citet{huang2023advancing}, for a brief review of the state of this research area.

\textbf{Memory entry types.} Transformer-XL \cite{dai-etal-2019-transformer} and MART \cite{lei-etal-2020-mart} use a cache to memorize the contextualized embeddings computed during the previous step at each layer. Compressive Transformer \cite{rae2019compressive} adds an additional compressed memory to collect the discarded entries from the Transformer-XL cache. Memorizing Transformer \cite{wu2021memorizing} and StreamingLLM \cite{xiao2023efficient} modify the transient Q/K/V structure of a dot-product attention, and allows K/Vs to persist. LongMem \cite{wang2023augmenting} is similar to Memorizing Transformer but further freezes the backbone LLM that generates K/Vs and introduces a separate trainable SideNet to overcome the staleness issue of Memorizing Transformer. Unlimiformer \cite{bertsch23neurips} stores the hidden states of encoder layers with no capacity limit, and retrieves top keys directly from the memory in cross attention after a reformulated attention equation. TRAMS \cite{yu2023trams} combines Transformer-XL with Unlimiformer, which retrieves top-$m$ keys from a memory pool size of $M$, which itself uses the FIFO policy despite the memory selection method for top keys. All of the above methods insert existing intermediate activations into the memory. Memformer \cite{wu-etal-2022-memformer} and TTM \cite{ryoo2023token} implicitly selects and creates, using a ``Write'' operation, memory entries from a combination of layer or model inputs and outputs and existing memory entries. \citet{liang2023unleashing} introduces a plain text memory outside the LLM, different from our activation memory that is inside the LLM. Our work extends Memorizing Transformer by memorizing not only the K/Vs but also the Qs, to support bidirectional attention over ``future'' K/Vs. Compared with the approaches that learn to compress or update the memory, our work follows the context length extension setup that takes an existing model and requires no further fine-tuning.

\textbf{Memory update methods.} Memorizing Transformer \cite{wu2021memorizing} and LongMem \cite{wang2023augmenting} both follow the FIFO update rule, like most prior sliding window attention methods, whereas StreamingLLM \cite{xiao2023efficient} defines attention sinks (a special memory partition) to keep initial positions. Since these models can only memorize predefined positions, they are difficult to adapt to attention methods with different score distributions or tasks that require memorization of other positions, or more generally the lost in the middle issue \cite{liu2023lost}. \citet{anagnostidis2023dynamic} propose to learn an adaptive attention mask to dynamically prune content from the autoregressive cache. Memory compression has also gained much attention. Compressive Transformer \cite{rae2019compressive} proposes various pooling and convolution methods to compress the Transformer-XL cache, where they use the ``most-used'' scheme, very similar to our LRA policy, as a baseline. Their results confirm that this method outperforms all other methods that do not need training. But the full potential of this scheme is underexplored. Expire-Span \cite{sukhbaatar2021not} learns the expiration for each activation. \citet{berchansky2023optimizing} filter positions with lowest attention scores at decoding time. ICAE \cite{ge2023context} uses an autoencoder to compress the context into a short sequence. MART \cite{lei-etal-2020-mart} uses a gated recurrent network to control the update of a memory. Memformer \cite{wu-etal-2022-memformer} uses an attention mechanism combining with a forget gate to update the memory. TTM \cite{ryoo2023token} uses learnable networks to create memory entries at each step. \citet{huang-hollenstein-2023-long} proposes to use eye tracking data to train a neural network to select and replace entries in the Transformer-XL cache. Most of these methods require additional training. In this paper, we leverage existing retrieval usage information to facilitate K/V memory update, similar to ``most-used'' scheme and require no training. We also acknowledge the complementary benefit of learnable memory operations, e.g. a learnable method to compress the ``least-used'' items may outperform any single compression method \citet{rae2019compressive}.

\textbf{Other uses of memory in LMs.} Besides the short-term memory in long context processing, long-term memory is also used in other scenarios to provide additional out-of-context knowledge. For example, \citet{borgeaud2021improving} uses a separate memory to store Wikipedia snippets to improve LM performance on other tasks. In this case, the memory is open for entries to insert before it becomes frozen in the downstream applications. \citet{de2023pre} proposes to retrieve augmentations in the precomputed passage memory for QA tasks. \citet{barraco2023little} uses a prototypical memory model to allow attention over activations obtained while processing other examples. While long term and short term memory share many design considerations in common, e.g. how to efficiently retrieve relevant entries from the memory using a trainable Q/K embedding, short term memory design in long context processing has additional challenges to satisfy the requirement of continuous insertion.

\textbf{Context length extension.} Researchers have studied the subproblem that tries to extend the context length of an existing LM pretrained using only relatively short sequences. Most work focuses on LMs using RoPE \cite{su2023roformer}, as it has become a popular choice in SOTA LLMs, but suffers from great extension difficulty \cite{kazemnejad2023impact, chen2023extending, peng2023yarn, han2023lm}. \citet{chen2023extending} proposes to use positional interpolation and \citet{peng2023yarn} proposes to modify the RoPE frequency to transform the new positions into the trained RoPE's ``comfort zone''. However, they both require an additional fine-tuning step to reach the desired level of performance. \citet{han2023lm} proposes the LM-infinite method that uses a predefined maximum ``allowed'' relative distance to cap the relative positions between tokens of a long sequence. This approach requires no fine-tuning and still achieves reasonable performance. In this paper, we follow the same context length extension setup, i.e. use existing pretrained models and directly apply them to long context problems without fine-tuning. We use a modified version of the LM-infinite method (detailed in Section \ref{sec:experiments}) when RoPE is used.

\section{\textsc{Attendre} Layer}

In this section, we first define the memory interface with eviction policies, and then describe how to combine memory modules to fulfill the wait-to-attend requirement.

\subsection{Memory \& Eviction Policies}

We identify two common use cases that require information of past steps and implement generic memory modules for the two types. The simpler use case is to memorize a single or a group of data (i.e. \textsc{insert} operation), and then use the data \emph{as a whole}, with no selection or filtering, at a future step (i.e. \textsc{getAll} operation). Transformer-XL \cite{dai-etal-2019-transformer} uses a data-only memory to cache the activations. We may also insert the auxiliary metadata (including positions, document ids, epochs, etc.) alongside the data, although we do not expect these data are used in indexing or retrieval.

The other use case is to provide an additional searchable key to accompany the values at the \textsc{insert}ion time, and then \textsc{retrieve} the most relevant keys and values (up to the size of the memory) using a query at a future step. Memorizing Transformer \cite{wu2021memorizing} uses a key-value memory to extend the K/V context. We may choose to reuse the keys and queries before or after the FFN sublayer or transformation (e.g. RoPE) inside the attention layer, or learn separately key and query networks independent of the attention layer. Likewise, the auxiliary metadata can be provided at insertion time and returned for the top retrieved keys and values. Moreover, intermediate computation results (including query-key similarity, masking, etc.) are also returned.

In both cases, a larger memory would allow to keep more context in past steps accessible by more subsequent steps, which can presumably improve the understanding of longer contexts. However, larger memory may not only increase the space complexity, but also sometimes the time complexity. As most prior works do, we limit the size of the memory. Further, we hypothesize that it may be suboptimal to evict the ``oldest'' entries. Instead, we propose to evict the least important entries, by employing widely adopted caching evicting policies, such as FIFO, LRU and LFU.

\begin{itemize}
    \item \textbf{FIFO (First-In First-Out).} Equivalent to a queue data structure, that evicts the ``oldest'' entries at insertion time. Sliding window attention \cite{beltagy2020longformer} and rolling K/V cache \cite{xiao2023efficient} are implementations of the FIFO policy. We use FIFO when we do not have the usage information and/or we require the entries should be evicted in the chronological order.
    
    \item \textbf{LRU (Least Recently Used).} A K/V is considered ``used'' if it is retrieved at least once by any query in the chunk, and all the K/Vs in the memory are sorted by the position of the query that last accessed them. ``Long-forgotten'' K/Vs are evicted.
    
    \item \textbf{LFU (Least Frequently Used).} We may further aggregate the usage statistics across batches by adding up the number of times each K/V is used in all previous steps. Those with the smallest usage counts are evicted.
\end{itemize}

For LRU and LFU, we keep track of the usage at the token level for both multi-head attention or multi-query attention \cite{shazeer2019fast}. In the case of multi-head attention, a K/V is considered ``used` if any head of the K/V is used, and all heads are evicted from the memory if the token is evicted\footnote{We can also have another variant that checks and evicts for each token-head separately. In our preliminary experiments, multi-query attention performs as well as multi-head attention in several long context tasks.}. There are two major issues with LRU and LFU: (1) the policy implicitly depends on the number of K/Vs to retrieve (a hyperparameter), and (2) the policy may assign the same priority to multiple K/Vs that are retrieved by the same query, and when needed, has to randomly evict K/Vs. To overcome these issues, we use LRA and LFA, which directly use the continuous attention score between the queries and the keys, after any transformation or bias, such as position bias or LM-Infinite, to replace the boolean use-or-not value. This is similar to the ``most-used'' baseline used in Compressive Transformer \cite{rae2019compressive}. Recent work on attention sorting \cite{peysakhovich2023attention} also suggests attention score may be a helpful indicator of document-level importance. We further use it at the token level to determine the importance of the corresponding activation.

\begin{itemize}
    \item \textbf{LRA (Least Recently Attended).} We compute the score for each K/V by aggregating attention score across all valid queries in the chunk (defined by the query-key 2-d mask), and also all heads in the multi-head attention case. We consider to use the attention score between only the last valid query position and the K/Vs (denoted as \textbf{LRA\textsubscript{last}}), or apply a max or sum pooling over all query positions (denoted as \textbf{LRA\textsubscript{max}} and \textbf{LRA\textsubscript{sum}} respectively). We note that both LRA\textsubscript{max} and LRA\textsubscript{sum} depend on the chunk size.
    
    \item \textbf{LFA (Least Frequently Attended).} We use the usage statistics across the steps in LFA, making this policy independent of the chunk size. We first optionally apply an exponential decay to the pairwise key-query attention score, i.e. $\exp \lambda ( i-i_\textrm{\textsubscript{max}} )$, where $\lambda$ is the decay factor, $i$ is the query position and $i_\textrm{\textsubscript{max}}$ is the maximum query position seen so far, and a similar decay to the aggregated per-K/V score from past steps, i.e. $\exp \lambda ( i'_\textrm{\textsubscript{max}}-i_\textrm{\textsubscript{max}} )$, where $i'_\textrm{\textsubscript{max}}$ is the maximum query position seen in the previous step. Then, we sum up the decayed attention scores of the K/Vs within and across chunks in all steps. We denote it as \textbf{LFA-$\lambda$}. When $\lambda > 0$, the policy can bias towards recently attended K/Vs.
\end{itemize}

Similar to other eviction policies, we need to specify the initial score for new positions. A high initial score can keep newly inserted positions in the memory for at least one step and guarantee to use them (by being attendees) in the immediate next step, but it may also result in passive eviction of important historic positions due to the capacity limit. We should set the initial score based on the score distribution, which often changes with the type of the position bias, the type and the location of the layer normalization and the depth of the layer. In our preliminary experiments, we found that setting the initial score to 1 to 2 standard deviations ($\sigma$) below the mean score ($\mu$) often leads to best performance. This way, the policy can keep more important positions and evict less important positions. The initial score also depends on the pooling method. Generally, a policy that assigns higher weights to recent positions prefers lower initial score, since these recent positions tend to better predict which K/Vs will remain important for future queries, and even the marginally important K/Vs can be more crucial than a blindly inserted new K/V on average. We use $\mu - \sigma$ as the initial score unless otherwise noted. Moreover, as the memory capacity increases, the model performance becomes less sensitive to the initial score, due to a decreased eviction ratio\footnote{We also tried other normalization methods alone or combined, e.g. clipping or softmax, and saw similar results.}.

The evicted entries are returned from the \textsc{insert} operation. We summarize the memory types with the supported operations in Table~\ref{tab:memory-types}.

\begin{table}[t]
\centering
\caption{Memory Types and Operations}
\label{tab:memory-types}
\begin{tabular}{l}
\hline
\multicolumn{1}{c}{\bf\tt DataOnlyMemory} \\
\hline
{\tt -evictionPolicy: EvictionPolicy} \\
\hline
{\tt +insert(value: V): EvictedD} \\
{\tt +getAll(): D} \\
\hline
\hline
\multicolumn{1}{c}{\bf\tt KeyValueMemory} \\
\hline
{\tt -evictionPolicy: EvictionPolicy} \\
\hline
{\tt +insert(key: K, value: V): EvictedKV} \\
{\tt +retrieve(query: Q): RetrievedKV} \\
\hline
\end{tabular}
\end{table}

\subsection{\textsc{Attendre} Layer}
\label{sec:wait-to-attend}

\begin{figure}[t]
    \centering
    \begin{subfigure}{\linewidth}
        \centering
        \includegraphics{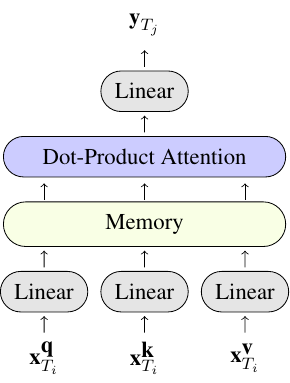}
        \caption{}
    \end{subfigure}
    \begin{subfigure}{\linewidth}
        \centering
        \includegraphics{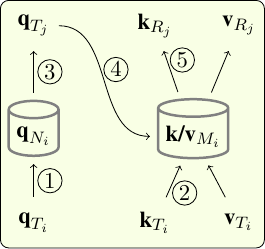}
        \caption{}
    \end{subfigure}
    \caption{\textsc{Attendre} layer with Q and KV memory storages. \textbf{(a)} The memory module is used to cache the linear transformed Q/K/V and in turn prepare time-shifted counterparts for dot-product attention. \textbf{(b)} \circled{1} Insert the query chunk $\textbf{q}_{T_i}$ into the Q memory. \circled{2} Insert the K/V chunks $\textbf{k}_{T_i}$ and $\textbf{v}_{T_i}$ into the K/V memory. \circled{3} Obtain the evicted query chunk $\textbf{q}_{T_j}$. \circled{4} Use the evicted query chunk $\textbf{q}_{T_j}$ to retrieve the K/V memory. \circled{5} Obtain top K/Vs $\textbf{k}_{R_j}$ and $\textbf{v}_{R_j}$.}
    \label{fig:wait-to-attend}
\end{figure}

\begin{figure*}[t]
    \centering
    \includegraphics{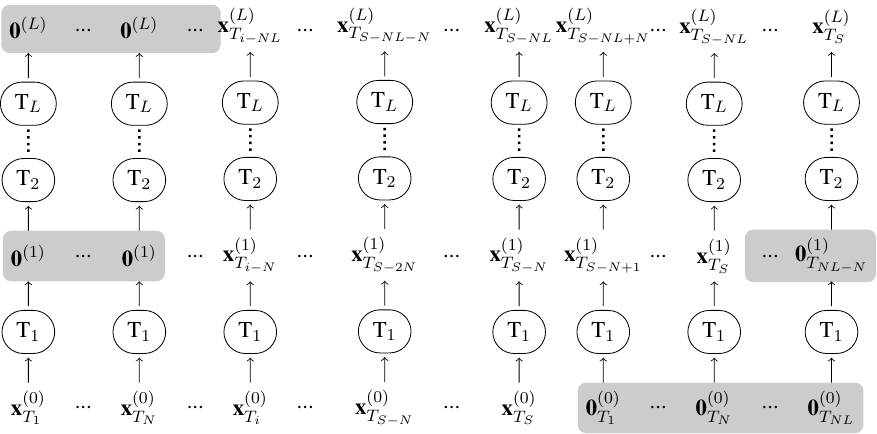}
    \caption{Time-shifted Transformer stack with a wait-to-attend layer inside the self-attention layer of each Transformer layer. The output of each layer shifts the input by the size of the Q memory $N$. The final output of a Transformer stack consisting of $L$ layers shifts the original input by $NL$. We postpad the input by $NL$ to ``drain'' the $L$ Q memory storages and trim the $NL$ paddings prepending the output sequence.}
    \label{fig:time-shifted-stack}
\end{figure*}

When inserting a stream of chunk sequences into a model that contains a memory module in a sequential order and reading them at the same time, we implicitly follow the \emph{causal order} or \emph{temporal order} between the chunks, i.e. each chunk can access all past chunks up to the current step but no future chunks. We only decide whether to allow the bidirectional dependency between positions within the chunk. This means when we use the memory inside an attention layer, we should assume the attention is \emph{unidirectional}, not \emph{bidirectional}.

An encoder architecture or the encoder tower in the encoder-decoder architecture expects only bidirectional inputs. Also, while a Transformer LM pretrained using PrefixLM \cite{raffel2020exploring} or UL2 \cite{tay2022ul2} objectives has generally gotten used to a mixture of unidirectional and bidirectional inputs during training, bidirectional information, if available, should continue to help contextualize in a decoder-only architecture. This motivates us to develop a key-value memory that can retrieve those that are inserted not only up to the step at which the query needs to retrieve, but also beyond the step!

We introduce the \textsc{Attendre} layer, a wait-to-attend mechanism with two memory modules: one data-only Q memory to ``delay'' queries and another key-value memory for K/Vs (Figure~\ref{fig:wait-to-attend}). We denote the sliced positions at step $i$ as $T_i$, and the corresponding linear transformed query, key, and value as $\textbf{q}_{T_i}$, $\textbf{k}_{T_i}$ and $\textbf{v}_{T_i}$ respectively. In the vanilla dot-product attention architecture, the output $o_s$ for each $s \in T_i$ is computed by

$$
\textbf{o}_s = \sum_{s' \in T_i} \frac{ \textrm{sim}(\textbf{q}_s, \textbf{k}_{s'}) }{ \sum_{s'' \in T_i} \textrm{sim}(\textbf{q}_s, \textbf{k}_{s''}) } \textbf{v}_{s'}
$$

With the \textsc{Attendre} layer, we first insert them into the Q memory and the K/V memory (\circled{1} and \circled{2}). Now, we obtain the updated Q memory and the K/V memory, which now (at step $i$) preserve indices $N_i$ ($\supseteq T_i$ before eviction) and $M_i$ ($\supseteq T_i$ before eviction). We further assume the Q memory uses FIFO policy to ensure ordered eviction\footnote{We can also try to use other eviction policy here if we consider some queries might require more ``future'' positions than others and hence stay in the memory for longer.}, and then the evicted query is exactly the same query that is inserted at a past step $j$ ($< i$) (\circled{3}). Instead of using the fresh query $\textbf{q}_{T_i}$ to retrieve the K/V memory as Memorizing Transformer \cite{wu2021memorizing}, we use the evicted query $\textbf{q}_{T_j}$ to retrieve the K/V memory, which has contained positions beyond $T_j$ (\circled{4}). We retrieve top-$K$ K/Vs for each query, whose position set $R_j$ ($\subseteq M_i$) has two trailing dimensions of query and retrieval instead of one trailing dimension of K/V, and contains positions beyond $T_j$ (\circled{5}). Finally, when using $\textbf{q}_{T_j}$ together with $\textbf{k}_{R_j}$ and $\textbf{v}_{R_j}$ in the subsequent dot-product attention, $\textbf{q}_{T_j}$ is allowed to attend to up to $N=|N_i|$ future K/V positions. Specifically, for each evicted query position $s \in T_j$, we compute its contextualized embedding using the modified attention formula:

$$
\textbf{o}_s = \sum_{1 \le r \le K} \frac{ \textrm{sim}(\textbf{q}_s, \textbf{k}_{sr}) }{ \sum_{1 \le r' \le K} \textrm{sim}(\textbf{q}_s, \textbf{k}_{sr'}) } \textbf{v}_{sr}
$$

\noindent where 2-d indices $sr$ and $sr' \in R_j$.

The idea of limiting the number of retrieved K/Vs are also used in other long sequence modeling works \cite{han2023lm, tworkowski2023focused, bertsch23neurips, yu2023trams}. We confirmed that retrieving fewer K/Vs reduces the physical memory usage of subsequent computation with little impact on the performance (up to a certain threshold)\footnote{We found the actual time complexity may increase with our current implementation on Cloud TPU. In fact, with the introduction of the two dimensional $R_j$ index, the dot product is now compiled into a loop fusion instead of a convolution fusion, where the latter has been optimized for the accelerator.}. Depending on the low-level library implementation, the top-K retrieval itself may at worst require element-wise distance calculation with each memory entry and a complete sorting step, and hence increases the overall computational cost. Similar to \citet{wu2021memorizing}, we also use the approximate top-k algorithm, and in Section \ref{sec:experiments}, we use a fixed value for this hyperparameter for fair comparison.

The size of the Q memory $N$ should be smaller than the size of the K/V memory, to avoid the evicted queries having to attend to only (in the FIFO K/V memory case) or mostly (in other cases) ``future'' K/Vs due to the K/V memory eviction, with no past or even ``present'' K/Vs (those that are generated for the same inputs as the evicted queries). When setting the size of the Q memory $N$ to half the size of the K/V memory, each evicted query can attend to roughly equal number of past and future K/Vs, which often yields a good result in our preliminary experiments.

We note that each \textsc{Attendre} layer ``delays'' (or right-shifts) the input by $N$ positions. A Transformer stack consisting of $L$ Transformer layers shifts the final output by $NL$ positions if all of them have an \textsc{Attendre} layer. It means, in order to complete the encoding of the prefixes in a decoder architecture, or the inputs in an encoder architecture, we need to either \textbf{flush} or \textbf{drain} the memory, before the autoregressive decoding begins, which does not require the Q memory due to its unidirectional nature. When we flush the memory, we introduce a single additional step where the entire residual queries in the Q memory take the place of the evicted queries for subsequent computation (via the \textsc{getAll} operation). We expect a higher than usual peak space usage during the flush ($N$ vs $S$ the chunk sequence length). Alternatively and more space-affordably, we can also perform the draining of the memory, which inserts padding chunks at each additional step, and obtains an evicted query of size $S$. We require $N$ padding tokens to drain one layer and $NL$ padding tokens to drain the entire stack. This value can still be small compared to the length of a long context. We illustrate the time-shifted Transformer stack architecture in Figure~\ref{fig:time-shifted-stack}.

In practice, we also reuse the similarity computed between $\textbf{q}_{T_j}$ and $\textbf{k}_{M_i}$ for the subsequent attention score computation if the same similarity distance is specified.
 
\textbf{Complexity.} We assume the sequence length of a chunk is $S$, the total number of chunks of the original long context input is $C$, the time complexity of processing the entire sequence in one step without chunking is $(CS)^2$ per layer using the vanilla Transformer architecture with the dot-product attention. If we feed each chunk into the model sequentially, the time complexity of processing the entire sequence is $CS^2$ per layer. We use $N=|N_i|$ and $M=|M_i|$ to denote the constant size of the Q memory and the K/V memory at any given step $i$, the time complexity becomes $(CS+N)M$ when we retrieve the entire $M$ K/V candidates, where $N$ padding positions are added to the original $CS$ positions for draining the Q memory. This can be further improved to $(CS+N)K$ when we can efficiently retrieve $K$ K/Vs for each query. We note that when $C$ is large, increasing the K/V memory size $M$ will increase the time complexity much more than increasing the Q memory size $N$. Therefore, our priority is to minimize the K/V memory size by using an evicting policy.

\subsection{Encoder Output Memory for Encoder-Decoder Architecture}

\begin{figure}[t]
    \centering
    \includegraphics{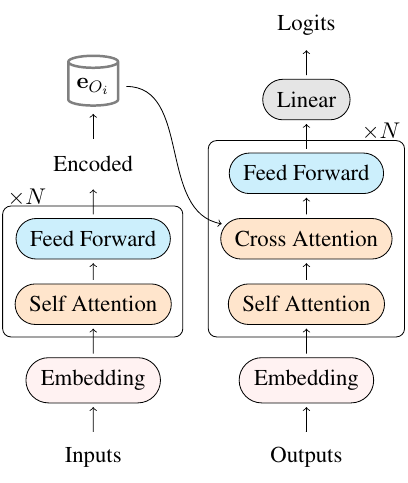}
    \caption{Encoder-decoder architecture with the additional encoder output memory $\textbf{e}$ to collect the outputs from the encoder. Each decoder layer uses the K/Vs from the encoder output memory $\textbf{e}$ instead to compute the cross attention.}
    \label{fig:encdec-architecture}
\end{figure}

Most previous works on memory-based long context modeling have focused on the decoder-only architecture, due to its simplicity and capability in performing various understanding and generation tasks. On the other hand, the encoder-decoder architecture has rarely been equipped with a memory attention to support understanding of contexts beyond a single chunk. The issue is that, although the encoder outputs in the subsequent steps get encoded using the contexts in the memory from earlier steps, they are still contextualized representations of their corresponding input positions after all. The encoder in the encoder-decoder architecture is not tasked to compress the input, unless a separate learning objective is used. For example, ICAE \cite{ge2023context} employs an autoencoder learning objective to summarize the input. \citet{al2022global} proposes a chunk selector network over the entire encoder output memory. Memformer \cite{wu-etal-2022-memformer} uses the class token with a memory writing network to update a fixed sized encoder memory iteratively. All of these methods require model training. Without these techniques, Unlimiformer \cite{bertsch23neurips} and many retrieval-augmented models choose to keep all possibly arbitrarily long encoder outputs.

Similar to Unlimiformer \cite{bertsch23neurips}, we apply a simple modification to the architecture by introducing an additional data-only memory medium \emph{encoder output memory} \textbf{e} (Figure~\ref{fig:encdec-architecture}). Since the decoding always occurs after the encoding in the encoder-decoder architecture, all the encoded inputs are inserted into \textbf{e} before they get used in the decoder stage. We cannot use the usage information from the decoder side to decide which entries to evict at insertion time. Therefore, we use the FIFO policy and always use the entire encoder output memory during the cross attention stage\footnote{It might also be helpful to borrow the usage information from the last encoder self attention layer to determine the importance of input positions. However, we are uncertain whether a position that is important to encoding is equally important to decoding.}. We note that the size of the encoder output memory only affects the computational complexity of decoding. Compared with Unlimiformer \cite{bertsch23neurips}, we define the encoder output memory separately from the K/V memory due to their different purposes, which allows individual customization and optimization.

\section{Experiment: Context Length Extension on TriviaQA}
\label{sec:experiments}

We use two pretrained models, which have been trained using texts \textbf{truncated to moderate sequence length} and \textbf{no memory}. We directly run zero-shot inference using these models with \textbf{no fine-tuning}. We use 1,000 examples from the unfiltered TriviaQA \cite{joshi-etal-2017-triviaqa} validation set, and report the mean exact match (EM) score. We use the following template to generate input:

\begin{verbatim}
    Question: {question}\n\n
    Context: {context}\n\n
    Answer: 
\end{verbatim}

\noindent where the context is concatenated from all relevant Wikipedia passages and search results. We choose to place the question only before the document, not both before and after, as in \citet{liu2023lost}. While the order of question and context in the template makes little difference to the performance of a model taking the entire input in a single step in theory\footnote{\citet{liu2023lost} found that the order of question and context also matters in practice due to other factors, including positional embeddings and/or pretraining objectives.}, it does affect the performance of a memory model that splits the input into chunks. In fact, the answer often occurs multiple times in the given context, including the last chunk, in the TriviaQA case. Therefore, if the question comes after the context, the model, when reading the last chunk, does not need to refer to the prior context to look for the question and the answer bearing context. When the question comes before the context, the model without an effective memory cannot recall the question that occurs in the first chunk. In general, placing the question before the context requires the eviction policy to keep the question in the memory indefinitely, and allows the memory model to select other relevant entries as it goes. On the other hand, placing the question after the context (i.e. no query-aware contextualization) challenges the memory to summarize the context solely based on its ``self'' (context to context) attention scores.

We use two pretrained and FLAN \cite{chung2022scaling}-tuned language models to perform this reading comprehension task:

\begin{itemize}
    \item \textbf{PaLM 2-S} \cite{anil2023palm} is the smallest model of the PaLM 2 trio, a family of Transformer-based models trained using a mixture of objectives. Since our goal is long context based reading comprehension, instead of closed-book QA, we report the result on PaLM 2-S, as opposed to other larger models as in \citet{anil2023palm}. The PaLM 2-S model is trained using moderate context length.
    
    \item \textbf{FLAN-T5 XXL} \cite{chung2022scaling} uses the encoder-decoder Transformer architecture and has 11B parameters, which is trained using span corruption objective \cite{raffel2020exploring} before further tuned using FLAN instruction dataset. The model is trained using inputs and targets both truncated after 512 tokens, as well as the traditional table-based relative position embedding instead of RoPE.
\end{itemize}

We use a variant of the LM-Infinite technique \cite{han2023lm} to extend context window of an existing LM pretrained with RoPE, since in our preliminary experiments, we found LM-Infinite performs better in the zero fine-tuning setup than other alternatives, including positional interpolation \cite{chen2023extending}, Yarn and NTK variants \cite{peng2023yarn}. In particular, we set $n_\textrm{local}$ to the actual context length used during pretraining\footnote{If unknown, we may also observe the performance gap between consecutive context length buckets to identify the effective context length.}. In contrast to the original LM-Infinite method, we do not specify the $n_\textrm{global}$ hyperparameter nor mask out the query-key pairs between the global and local boundaries. Instead, we compute all pairs outside the local diagonal band using the global relative distance, which allows us to preserve as much context information as possible, and let the eviction and retrieval steps to decide which to exclude.

We compare our proposed methods LRA\textsubscript{last}, LRA\textsubscript{max}, LRA\textsubscript{sum}, LFA-0, LFA-$10^{-3}$, with the Memorizing Transformer (i.e. the FIFO policy, MT) \cite{wu2021memorizing} and StreamingLLM (i.e. the attention sink, AS) \cite{xiao2023efficient} baselines, where we set the sink size to 4. When combining LM-Infinite with Attention Sink, we use original positions (instead of in-cache positions) for the local branch and assign fixed zero-max positions for the global branch.

\subsection{Results on PaLM 2-S}

\begin{table}
\centering
\caption{Results using the original PaLM 2-S model. We vary the sequence length (S, top row) and report the mean EM score (bottom row).}
\label{tab:results-palm2-original}
\begin{tabular}{ccccccc}
\hline
128  & 256    & 512    & 1024   & 2048   & 4196    \\
\hline
2.26 & 9.62   & 24.24  & 39.94  & 53.88  & 59.27   \\
\hline
\end{tabular}
\end{table}

\begin{table*}
\centering
\caption{Experimental results using PaLM 2-S model with memory. We vary the K/V memory size (M) and the Q memory size (N), and report the mean EM score. We compare with Memorizing Transformer (MT) \cite{wu2021memorizing} and Attention Sink (AS) \cite{xiao2023efficient} policies. We fixed the chunk size to 128 and retrieved top 128 K/Vs from the K/V memory.}
\label{tab:results-palm2-memory}
\begin{tabular}{cc|cccccccc}
\hline
M     & N    & MT     & AS     & LRA\textsubscript{last}  & LRA\textsubscript{max}  & LRA\textsubscript{sum}  & LFA-0  & LFA-$10^{-3}$ \\
\hline
128   & 0    & 2.26   & 11.38  & 42.69     & 42.89    & 47.60    & \textbf{50.74}  & 47.79 \\
256   & 0    & 8.54   & 17.47  & \textbf{51.91}     & 46.42    & 49.66    & 50.25  & 50.54 \\
512   & 0    & 21.69  & 26.79  & 51.13     & 48.58    & 50.83    & 51.23  & \textbf{51.62} \\
1024  & 0    & 37.00  & 40.33  & 51.82     & 50.54    & 52.70    & \textbf{52.99}  & 51.72 \\
2048  & 0    & 48.87  & 50.54  & 52.50     & 51.62    & 52.21    & \textbf{52.60}  & 51.72 \\
\hline
128   & 128  & 6.67   & 13.25  & \textbf{55.74}     & 50.83    & 54.27    & 51.91  & 54.17 \\
512   & 256  & 19.04  & 23.65  & 57.31     & 56.92    & 58.68    & 54.47  & \textbf{58.88} \\
1024  & 512  & 33.95  & 37.10  & 61.14     & \textbf{63.40}    & 62.61    & 58.49  & 62.22 \\
2048  & 1024 & 52.21  & 52.31  & 66.63     & \textbf{69.38}    & 67.62    & 63.00  & 68.99 \\
\hline
\end{tabular}
\end{table*}

We first report the results using the original PaLM 2-S model with no memory in Table~\ref{tab:results-palm2-original}, where we vary the input sequence length, i.e. the chunk size $S$, which is equivalent to reading only the last $S$ tokens due to the absence of a memory. Unsurprisingly, we see the model performs poorly when we set the sequence length to 128 and starts to improve as we increase the chunk size. Then, we report the results when varying the memory size in Table~\ref{tab:results-palm2-memory}. In our experiment, we fixed the chunk size to 128, and retrieved only the top 128 K/Vs from the K/V memory, instead of all the K/Vs for fair comparison between different $M$ values and a reduced physical memory usage. We use $\mu - 2\sigma$ as the initial score for LRA\textsubscript{last} and $\mu - 1.5\sigma$ for LRA\textsubscript{max} and LRA\textsubscript{sum}, since we found these values help improve the performance with the PaLM 2 model.

Our first observation is that even when we set $M$ to 128, i.e. we keep the attendee pool size identical to the vanilla attention mechanism, the LRA and LFA policies can still achieve reasonable performance (up to 50.74), much higher than the MT and the AS baselines, and close to the original PaLM 2-S model when the sequence length is 2,048 (53.88). The AS policy, which keeps the four initial positions, including the initial two tokens in the question, helps it outperform the MT. When we further increase M to 2,048, we see the performance of the baseline methods MT and AS greatly improves ($+2,062.39\%$ and $+344.11\%$ respectively), suggesting the benefit of additional attendees in the extended context windows. However, the performance of our methods only marginally improves ($+4.4\%$), suggesting that the LRA and LFA policies can effectively select positions to keep in the memory even when the memory size is small. We also see that our methods continue to outperform the baselines, confirming the effectiveness of the proposed automatically adaptive LRA and LFA policies. Then, we add the Q memory and set the size of the Q memory ($N$) to half the size of the K/V memory ($M$). We see that, in contrast to the baseline methods, our methods continue to benefit from the additional Q memory. With 1024 K/V memory slots and 512 Q memory slots, the LFA\textsubscript{max} policy enables the PaLM 2-S model, using a chunk size of 128, to achieve an average EM of 63.40, higher than the EM obtained by the original PaLM 2-S model using an input length of 4,196 (59.27).

Among the LRA policy variants, we see that the LRA\textsubscript{last} policy and the LRA\textsubscript{sum} policy outperform the LRA\textsubscript{max} policy when the K/V memory size is small ($\le +384$) or no Q memory is used, but underperforms the LRA\textsubscript{max} policy otherwise, suggesting that a large memory expects the policy to preserve not only the universally ``popular'' positions, but also those that are very important to a minority of queries. When we further allow aggregating the usage statistics across chunks, we see that the LFA-$10^{-3}$ policy performs better than the LFA-0 policy when the Q memory is used, and achieves the best or near best EM score ($\le -2.82\%$ from the best). Without the Q memory, the performance difference between the two policies is much smaller. Our hypothesis is that, as we increase the size of the Q memory, the query evicted from the Q memory starts to lag behind the last inserted K/Vs, and has become less predictive of the importance of each K/V to future queries. Among these query positions, those at the tail are relatively recent and thus their attention scores have better predictive power, whose weights can be boosted by the LFA-$10^{-3}$ policy alongside LRA\textsubscript{last} and LRA\textsubscript{sum}.

\subsection{Results on FLAN-T5 XXL}

\begin{table}
\centering
\caption{Results using the original FLAN-T5 XXL model. We vary the sequence length (S, top row) and report the mean EM score (bottom row).}
\label{tab:results-t5-original}
\begin{tabular}{ccccccc}
\hline
128  & 256    & 512    & 1024   & 2048   & 4196    \\
\hline
3.53 & 8.24   & 17.86  & 31.70  & 48.77  & 57.80   \\
\hline
\end{tabular}
\end{table}

\begin{table*}
\centering
\caption{Experimental results using FLAN-T5 XXL with memory. We vary the K/V memory size (M) and the Q memory size (N), and report the mean EM score. We compare with Memorizing Transformer (MT) \cite{wu2021memorizing} and Attention Sink (AS) \cite{xiao2023efficient} policies. We fixed the chunk size to 128 and retrieved top 128 K/Vs from the memory. We set the size of encoder output memory (O) to 8,192.}
\label{tab:results-t5-memory}
\begin{tabular}{cc|ccccccc}
\hline
M     & N    & MT     & AS     & LRA\textsubscript{last}  & LRA\textsubscript{max}  & LRA\textsubscript{sum}  & LFA-0  & LFA-$10^{-3}$ \\
\hline
128   & 0    & 25.42  & 37.78  & 42.00     & 40.33    & 41.41    & 32.48  & \textbf{42.20} \\
256   & 0    & 31.70  & 39.25  & 44.06     & 40.92    & \textbf{45.34}    & 40.53  & 44.95 \\
512   & 0    & 36.31  & 41.90  & 44.85     & 43.47    & 45.93    & 44.06  & \textbf{46.91} \\
1024  & 0    & 40.14  & 43.18  & 47.20     & 45.63    & 47.69    & 44.75  & \textbf{48.38} \\
2048  & 0    & 45.24  & 46.61  & 47.30     & 47.20    & 47.01    & 46.03  & \textbf{48.48} \\
\hline
256   & 128  & 30.32  & 38.57  & 47.89     & 46.61    & \textbf{48.68}    & 41.12  & 48.48 \\
512   & 256  & 42.00  & 47.30  & 51.82     & 50.15    & \textbf{53.39}    & 47.89  & 52.50 \\
1024  & 512  & 52.80  & 50.44  & 56.92     & 54.76    & \textbf{58.29}    & 53.29  & 57.02 \\
2048  & 1024 & 54.47  & 56.62  & 56.53     & 57.61    & \textbf{58.39}    & 56.62  & 57.90 \\
\hline
\end{tabular}
\end{table*}

We report the results using the original FLAN-T5 XXL model in Table~\ref{tab:results-t5-original}. It shows that although the model is trained using only inputs truncated at 512 tokens and has not been ``patched'' using any LM-Infinite like method, it can process much longer inputs and achieve results only slightly worse than the PaLM 2-S model ($-2.4\%$ when the sequence length is set to 4,196).

We also report the results using memory enabled FLAN-T5 XXL model variants in Table~\ref{tab:results-t5-memory}, where we also fix the chunk size to 128 and retrieve top 128 K/Vs from the memory. We set the size of the encoder output memory to 8,192 throughout the experiment, and focus our discussion on the K/V and Q memory designs. We may consider to improve the memory usage in a future work. From Table~\ref{tab:results-t5-memory}, we see that the results using the FLAN-T5 XXL model have several similarities to those using the PaLM 2-S model: (1) our proposed methods mostly outperforms the baseline methods, especially when the memory size ($M$) is small, and (2) increasing the K/V memory size improves the performance across all methods, more with the baseline methods ($+77.98\%$ and $+28.95\%$) and less with the LRA and LFA policies (as little as $+14.88\%$), and (3) adding the Q memory and setting the size to half the size of the K/V memory also improves the performance of our methods and most baseline methods. When setting $M$ to 1,024 and $N$ to 512, the LRA\textsubscript{sum} achieves an EM of 58.29, outperforming the original FLAN-T5 XXL model when the sequence length is set to 4,096 (57.80), resulting a huge efficiency improvement.

We also observed that the performance gap between the baseline methods and our methods in this experiment is smaller than that in the PaLM 2-S experiment, thanks to the large encoder output memory. In fact, if we compare the first column in Table~\ref{tab:results-t5-original} (3.53) and the first row and the first column in Table~\ref{tab:results-t5-memory} (25.42), the only difference between the two setups is the size of the encoder output memory (effectively 128 vs 8,192). Therefore, in contrast to the PaLM 2-S experiments, some encoded positions, though not fully contextualized due to the less effective eviction policies of the baseline methods, still preserve some useful information, and reside inside the large encoder output memory, when the size of other memories is small. This also explains the rapid ``catch-up'' of the baseline methods with a large $M$, a threshold needed for local contextualization, since the long range dependency for accurate answer decoding is again taken care of by the large encoder output memory. The PaLM 2 model does not enjoy this ``convenience''.

Among all LRA and LFA policies, we first note that the LFA-0 policy is consistently the lowest performer, which differs from other policies in that it assigns the same weight to all the query positions, including the initial positions in the first chunk. The best results are often obtained by either the LRA\textsubscript{sum} policy or the LFA-$10^{-3}$ policy, which both assign higher weights to recent attention scores and lower weights to past attention scores.

\section{Conclusion}

In this paper, we propose to equip a memory module with eviction policies, such as LRA or LFA, to reduce the memory size and adapt to various architectures. We also propose the \textsc{Attendre} layer, a wait-to-attend mechanism by retrieving the key-value memory (K/V memory) with evicted queries in the query memory (Q memory) to support bidirectional attention. We evaluate the proposed methods using the TriviaQA task on two different Transformer-based pretrained models. We found that the proposed policies outperform the baseline methods. Also, we found a model trained using only bidirectional attention does require the proposed the \textsc{Attendre} layer to reach the performance of the original model that takes the long sequence at once.

An immediate next step is to evaluate and observe the performance on a wider range of tasks, as we noted in the paper, the TriviaQA task may not require a very long range dependency, even for long inputs. We will also be interested in fine-tuning a model with the memory modules, which includes updating the parameters of the original backbone model and/or updating a set of memory specific parameters, making them adaptable to other tasks. One challenge is that the Q memory stops the backward propagation of gradients, since the eviction of queries does not happen in the same step as the insertion and computation of these queries. However, as most modern LLMs are trained using \texttt{remat}\footnote{\url{https://jax.readthedocs.io/en/latest/jep/11830-new-remat-checkpoint.html}} \cite{chen2016training} to improve physical memory usage, we can further memorize and then provide the past evicted inputs that correspond to the gradients, instead of the newly inserted inputs, in the backward step. We would also like to compress the memory further. Fine-tuning the memory models for a small number of steps can also allow them to learn to contextualize from fewer positions, making it a possible solution to further compress the memory. Alternatively, we can employ a portable and learnable memory module, external to the original memory, that can compress the context. Last but not least, we need to consider to compress the encoder output memory for the encoder-decoder architecture, e.g. \citet{ge2023context}.

\bibliography{references}

\appendix

\end{document}